\newif\ifJOURNAL
\JOURNALfalse
\newif\ifCONF
\CONFfalse
\newif\ifarXiv
\arXivfalse
\newif\ifWP
\WPfalse
\newif\ifFULL
\FULLfalse
\newif\ifLATIN
\LATINfalse

\arXivtrue

\ifCONF\LATINtrue\fi	
\ifarXiv\LATINtrue\fi	

\newif\ifnotCONF	
\notCONFtrue
\ifCONF\notCONFfalse\fi

\newif\ifnotarXiv	
\notarXivtrue
\ifarXiv\notarXivfalse\fi

\newif\ifTR		
\TRfalse
\ifarXiv\TRtrue\fi
\ifWP\TRtrue\fi
\newif\ifnotTR
\notTRtrue
\ifarXiv\notTRfalse\fi
\ifWP\notTRfalse\fi

\newif\ifnotLATIN	
\notLATINtrue
\ifLATIN\notLATINfalse\fi

\ifCONF
  \newcommand{\DFI}{vovk/etal:2005AIStats-local}
  \newcommand{\DFII}{vovk/etal:2005ALT}
  \newcommand{\DFIII}{vovk:2005ALT-DF03}
  \newcommand{\DFIV}{vovk:2005ALT-DF04}
  \newcommand{\DFV}{vovk:2006China-local}
  \newcommand{\DFVI}{vovk:2006COLT-DF06-local}
  \newcommand{\DFVII}{vovk:2006COLT-DF07-local}
  \newcommand{\DFVIII}{DF08arXiv}
  \newcommand{\DFIX}{DF09arXiv-local}
  \newcommand{\DFX}{DF10arXiv-local}
  \newcommand{\AdamsFournier}{adams/fournier:2003}
  \newcommand{\VovkTCS}{vovk:2001-local}
\fi
\ifarXiv
  \newcommand{\DFI}{DF01arXiv}
  \newcommand{\DFII}{DF02arXiv}
  \newcommand{\DFIII}{DF03arXiv}
  \newcommand{\DFIV}{DF04arXiv}
  \newcommand{\DFV}{DF05arXiv}
  \newcommand{\DFVI}{DF06arXiv}
  \newcommand{\DFVII}{DF07arXiv}
  \newcommand{\DFVIII}{DF08arXiv}
  \newcommand{\DFIX}{DF09arXiv-local}
  \newcommand{\AdamsFournier}{adams/fournier:2003full}
  \newcommand{\VovkTCS}{vovk:2001}
\fi
\ifWP
  \newcommand{\DFI}{GTP8}
  \newcommand{\DFII}{GTP10}
  \newcommand{\DFIII}{GTP13}
  \newcommand{\DFIV}{GTP14}
  \newcommand{\DFV}{GTP11}
  \newcommand{\DFVI}{GTP16}
  \newcommand{\DFVII}{GTP17}
  \newcommand{\DFVIII}{DF08arXiv}
  \newcommand{\DFIX}{DF09arXiv}
  \newcommand{\AdamsFournier}{adams/fournier:2003full}
  \newcommand{\VovkTCS}{vovk:2001}
\fi
\ifFULL
  \newcommand{\DFI}{DF01arXiv}
  \newcommand{\DFII}{DF02arXiv}
  \newcommand{\DFIII}{DF03arXiv}
  \newcommand{\DFIV}{DF04arXiv}
  \newcommand{\DFV}{DF05arXiv}
  \newcommand{\DFVI}{DF06arXiv}
  \newcommand{\DFVII}{DF07arXiv}
  \newcommand{\DFVIII}{DF08arXiv}
  \newcommand{\DFIX}{DF09arXiv}
  \newcommand{\AdamsFournier}{adams/fournier:2003full}
  \newcommand{\VovkTCS}{vovk:2001}
\fi

\ifnotLATIN
  \newcommand{\KabanovEtAl}{kabanov/etal:1977}
\fi
\ifLATIN
  \newcommand{\KabanovEtAl}{kabanov/etal:1977latin}
\fi

\documentclass{article}
\usepackage{amsmath,amsfonts,amssymb,latexsym,graphicx}
\newcommand{\Extra}[1]{}

\ifWP
\documentclass[toc]{gtarticle}
\usepackage{amsmath,amsfonts,amssymb,latexsym,epsfig,graphicx,epigraph}
\renewcommand{\Extra}[1]{#1}
\fi

\ifFULL
\documentclass{article}
\usepackage{amsmath,amsfonts,amssymb,latexsym,color,epigraph,graphicx,epigraph}
\newcommand{\Extra}[1]{\red{#1}}
\newcommand{\red}[1]{\textcolor{red}{#1}}

\newcommand{\bluebegin}{\begingroup\color{blue}}
\newcommand{\blueend}{\endgroup}

\fi

\allowdisplaybreaks[3]

\ifnotLATIN
\input{OT2enc.def}

\fi

\newcommand{\Vladimir}{Vladimir}
\newcommand{\DOT}{.}

\newcommand{\st}{\mathrel{\!|\!}}
\newcommand{\Parallel}{\mathop{\parallel}}
\newcommand{\diam}{\mathop{\mathrm{diam}}}

\newcommand{\ccc}{\mathbf{c}}		
\newcommand{\FFF}{\mathcal{F}}		
\newcommand{\GGG}{\mathcal{G}}		
\newcommand{\KKK}{\mathbf{K}}		
\newcommand{\kkk}{\mathbf{k}}		

\ifnotCONF
\fi
\ifCONF
\fi

\newcommand{\Exp}{\mathop{\mathrm{Exp}}\nolimits}
\newcommand{\Span}{\mathop{\mathrm{span}}\nolimits}

\ifnotCONF
\newcommand{\bbbr}{\mathbb{R}}		
\newtheorem{lemma}{Lemma}
\newtheorem{proposition}{Proposition}
\newtheorem{corollary}{Corollary}
\newtheorem{theorem}{Theorem}
\newenvironment{proof}
  {\trivlist\item[\hskip\labelsep\textbf{Proof}]}
  {\endtrivlist}
\fi

\newcommand{\boxforqed}{\rule{.3em}{1.5ex}}
\newcommand{\qedtext}{\unskip\nobreak\hfil
  \penalty50\hskip1em\null\nobreak\hfil\boxforqed
  \parfillskip=0pt\finalhyphendemerits=0\endgraf}

\newenvironment{remark*}
  {\trivlist\item[\hskip\labelsep{\bfseries Remark}]\relax}
  {\endtrivlist}

\newlength{\IndentI}
\newlength{\IndentII}
\newlength{\IndentIII}
\newlength{\IndentIV}
\newlength{\WidthI}
\newlength{\WidthII}
\newlength{\WidthIII}
\newlength{\WidthIV}
\ifCONF
\title{Leading strategies\\in competitive on-line prediction}
\author{Vladimir Vovk}
\institute{Computer Learning Research Centre,
  Department of Computer Science\\
  Royal Holloway, University of London,
  Egham, Surrey TW20 0EX, UK\\
  \email{vovk@cs.rhul.ac.uk}}
\fi

\ifarXiv
\title{Leading strategies\\in competitive on-line prediction}
\author{Vladimir Vovk\\
\texttt{vovk{\rm@}cs.rhul.ac.uk}\\
\texttt{http://vovk.net}}
\fi

\ifWP
\title{Leading strategies\\in competitive on-line prediction}
\author{Vladimir Vovk}

\twodatestrue

\fi

\ifFULL
\title{Leading strategies\\in competitive on-line prediction}
\author{Vladimir Vovk\\
\texttt{vovk{\rm@}cs.rhul.ac.uk}\\
\texttt{http://vovk.net}}
\fi

\begin{document}
\maketitle
\begin{abstract}
  We start from a simple asymptotic result
  for the problem of on-line regression with the quadratic loss function:
  the class of continuous limited-memory prediction strategies
  admits a ``leading prediction strategy'',
  which not only asymptotically performs at least as well
  as any continuous limited-memory strategy
  but also satisfies the property
  that the excess loss of any continuous limited-memory strategy
  is determined by how closely it imitates the leading strategy.
  More specifically,
  for any class of prediction strategies
  constituting a reproducing kernel Hilbert space
  we construct a leading strategy,
  in the sense that the loss of any prediction strategy
  whose norm is not too large
  is determined by how closely it imitates the leading strategy.
  This result is extended to the loss functions
  given by Bregman divergences and by strictly proper scoring rules.
\end{abstract}


\section{Introduction}
\label{sec:introduction}

Suppose $\FFF$ is a normed function class of prediction strategies
(the ``benchmark class'').
It is well known that, under some restrictions on $\FFF$,
there exists a ``master prediction strategy''
(sometimes also called a ``universal strategy'') that performs
almost as well as the best strategies in $\FFF$ whose norm is not too large
(see, e.g., \cite{cesabianchi/long/warmuth:1996,auer/etal:2002}).
The ``leading prediction strategies'' constructed in this paper
satisfy a stronger property:
the loss of any prediction strategy in $\FFF$ whose norm is not too large
exceeds the loss of a leading strategy by the divergence
between the predictions output by the two prediction strategies.
Therefore,
the leading strategy implicitly serves as a standard
for prediction strategies $F$ in $\FFF$ whose norm is not too large:
such a prediction strategy $F$ suffers a small loss
to the degree that its predictions resemble the leading strategy's predictions,
and the only way to compete with the leading strategy is to imitate it.

We start the formal exposition with a simple asymptotic result
(Proposition \ref{prop:quadratic-asymptotic} in \S\ref{sec:regression})
asserting the existence of leading strategies
in the problem of on-line regression with the quadratic loss function
for the class of continuous limited-memory prediction strategies.
To state a non-asymptotic version of this result
(Proposition \ref{prop:quadratic-RKHS})
we introduce several general definitions
that are used throughout the paper.
In the following two sections
Proposition \ref{prop:quadratic-RKHS} is generalized in two directions,
to the loss functions given by Bregman divergences
(\S\ref{sec:bregman})
and by strictly proper scoring rules
(\S\ref{sec:proper}).
Competitive on-line prediction typically avoids making any stochastic assumptions
about the way the observations are generated,
but in \S\ref{sec:stochastic} we consider, mostly for comparison purposes,
the case where observations are generated stochastically.
That section contains most of the references to the related literature,
although there are bibliographical remarks scattered throughout the paper.
\ifnotCONF
  The proofs are gathered in \S\ref{sec:proofs}.
\fi
The final section, \S\ref{sec:conclusion},
discusses 
possible directions of further research.

There are many techniques for constructing master strategies,
such as gradient descent,
strong and weak aggregating algorithms,
following the perturbed leader,
defensive forecasting,
to mention just a few.
In this paper we will use defensive forecasting
(proposed in \cite{\DFIV} and based on \cite{\DFI,\DFIII}
and much earlier work by Levin, Foster, and Vohra).
The master strategies constructed using defensive forecasting
automatically satisfy the stronger properties required of leading strategies;
on the other hand,
it is not clear whether leading strategies can be constructed
using other techniques.

\section{On-line quadratic-loss regression}
\label{sec:regression}

Our general prediction protocol is:

\bigskip

\noindent
\textsc{On-line prediction protocol}\nopagebreak

\parshape=5
\IndentI  \WidthI
\IndentII \WidthII
\IndentII \WidthII
\IndentII \WidthII
\IndentI  \WidthI
\noindent
FOR $n=1,2,\dots$:\\
  Reality announces $x_n\in\mathbf{X}$.\\
  Predictor announces $\mu_n\in\mathbf{P}$.\\
  Reality announces $y_n\in\mathbf{Y}$.\\
END FOR.

\bigskip

\noindent
At the beginning of each round $n$ Forecaster is given some side information $x_n$
relevant to predicting the following observation $y_n$,
after which he announces his prediction $\mu_n$.
The side information is taken from the \emph{information space} $\mathbf{X}$,
the observations from the \emph{observation space} $\mathbf{Y}$,
and the predictions from the \emph{prediction space} $\mathbf{P}$.
The error of prediction is measured by a \emph{loss function}
$\lambda:\mathbf{Y}\times\mathbf{P}\to\bbbr$,
so that $\lambda(y_n,\mu_n)$ is the loss suffered by Predictor on round $n$.

A \emph{prediction strategy} is a strategy for Predictor in this protocol.
More explicitly, each prediction strategy $F$
maps each sequence
\begin{equation}\label{eq:s}
  s
  =
  (x_1,y_1,\dots,x_{n-1},y_{n-1},x_n)
  \in
  \mathbf{S}
  :=
  \bigcup_{n=1}^{\infty}
  \left(
    \mathbf{X}
    \times
    \mathbf{Y}
  \right)^{n-1}
  \times
  \mathbf{X}
\end{equation}
to a prediction $F(s)\in\bbbr$;
we will call $\mathbf{S}$ the \emph{situation space}
and its elements \emph{situations}.
We will sometimes use the notation
\begin{equation}\label{eq:s-n}
  s_n
  :=
  (x_1,y_1,\dots,x_{n-1},y_{n-1},x_n)
  \in
  \mathbf{S},
\end{equation}
where $x_i$ and $y_i$ are Reality's moves in the on-line prediction protocol.

In this section we will always assume
that $\mathbf{Y}=[-Y,Y]$ for some $Y>0$,
$[-Y,Y]\subseteq\mathbf{P}\subseteq\bbbr$,
and $\lambda(y,\mu)=(y-\mu)^2$;
in other words,
we will consider the problem of on-line quadratic-loss regression
(with the observations bounded in absolute value by a known constant $Y$).

\subsection*{Asymptotic result}

Let $k$ be a positive integer.
We say that a prediction strategy $F$ is \emph{order $k$ Markov}
if $F(s_n)$ depends on (\ref{eq:s-n}) only via
$x_{\max(1,n-k)},y_{\max(1,n-k)},\ldots,x_{n-1},y_{n-1},x_n$.
More explicitly, $F$ is order $k$ Markov if and only if
there exists a function
\begin{equation*}
  f:
  \left(
    \mathbf{X}
    \times
    \mathbf{Y}
  \right)^{k}
  \times
  \mathbf{X}
  \to
  \mathbf{P}
\end{equation*}
such that, for all $n>k$ and all (\ref{eq:s-n}),
\begin{equation*}
  F
  (s_n)
  =
  f
  (x_{n-k},y_{n-k},
  \ldots,
  x_{n-1},y_{n-1},x_n).
\end{equation*}
A \emph{limited-memory} prediction strategy
is a prediction strategy which is order $k$ Markov for some $k$.
(The expression ``Markov strategy'' being reserved for ``order 0 Markov strategy''.)
\begin{proposition}\label{prop:quadratic-asymptotic}
  Let $\mathbf{Y}=\mathbf{P}=[-Y,Y]$ and $\mathbf{X}$ be a metric compact.
  There exists a strategy for Predictor that guarantees
  \begin{equation}\label{eq:quadratic-asymptotic}
    \frac1N
    \sum_{n=1}^N
    \left(
      y_n-\mu_n
    \right)^2
    +
    \frac1N
    \sum_{n=1}^N
    \left(
      \mu_n-\phi_n
    \right)^2
    -
    \frac1N
    \sum_{n=1}^N
    \left(
      y_n-\phi_n
    \right)^2
    \to
    0
  \end{equation}
  as $N\to\infty$
  for the predictions $\phi_n$ output by any continuous limited-memory prediction strategy.
\end{proposition}
The strategy whose existence is asserted by Proposition \ref{prop:quadratic-asymptotic}
is a leading strategy
in the sense discussed in \S\ref{sec:introduction}:
the average loss of a continuous limited-memory strategy $F$
is determined by how well it manages to imitate
the leading strategy.
And once we know the predictions made by $F$ and by the leading strategy,
we can find the excess loss of $F$ over the leading strategy
without need to know the actual observations.

\subsection*{Leading strategies for reproducing kernel Hilbert spaces}

In this subsection we will state a non-asymptotic version
of Proposition  \ref{prop:quadratic-asymptotic}.
Since $\mathbf{P}=\bbbr$ is a vector space,
the sum of two prediction strategies
and the product of a scalar (i.e., real number) and a prediction strategy
can be defined pointwise:
\begin{equation*}
  (F_1+F_2)(s)
  :=
  F_1(s)+F_2(s),
  \quad
  (cF)(s)
  :=
  cF(s),
  \qquad
  s\in\mathbf{S}.
\end{equation*}
Let $\FFF$ be a Hilbert space of prediction strategies
(with the pointwise operations of addition and multiplication by scalar).
Its \emph{embedding constant} $\ccc_{\FFF}$ is defined by
\begin{equation}\label{eq:embedding-constant}
  \ccc_{\FFF}
  :=
  \sup_{s\in\mathbf{S}}
  \sup_{F\in\FFF:\left\|F\right\|_{\FFF}\le1}
  \left|F(s)\right|.
\end{equation}
We will be interested in the case $\ccc_{\FFF}<\infty$
and will refer to $\FFF$ satisfying this condition
as \emph{reproducing kernel Hilbert spaces (RKHS) with finite embedding constant}.
(More generally, $\FFF$ is said to be an \emph{RKHS}
if the internal supremum in (\ref{eq:embedding-constant})
is finite for each $s\in\mathbf{S}$.)
In our informal discussions we will be assuming
that $\ccc_{\FFF}$ is a moderately large constant.
\begin{proposition}\label{prop:quadratic-RKHS}
  Let $\mathbf{Y}=[-Y,Y]$, $\mathbf{P}=\bbbr$,
  and $\FFF$ be an RKHS of prediction strategies
  with finite embedding constant $\ccc_{\FFF}$.
  There exists a strategy for Predictor that guarantees
  \begin{multline}\label{eq:quadratic-RKHS}
    \left|
      \sum_{n=1}^N
      \left(
        y_n-\mu_n
      \right)^2
      +
      \sum_{n=1}^N
      \left(
        \mu_n-\phi_n
      \right)^2
      -
      \sum_{n=1}^N
      \left(
        y_n-\phi_n
      \right)^2
    \right|\\
    \le
    2Y
    \sqrt{\ccc_{\FFF}^2+1}
    \left(
      \left\|
        F
      \right\|_{\FFF}
      +
      Y
    \right)
    \sqrt{N},
    \qquad
    \forall N\in\{1,2,\ldots\}
    \enspace
    \forall F\in\FFF,
  \end{multline}
  where $\phi_n$ are $F$'s predictions,
  $
    \phi_n:=F(s_n)
  $.
\end{proposition}
For an $F$ whose norm is not too large
(i.e., $F$ satisfying $\left\|F\right\|_{\FFF}\ll N^{1/2}$),
(\ref{eq:quadratic-RKHS}) shows that
\begin{equation*}
  \frac{1}{N}
  \sum_{n=1}^N
  \left(
    y_n-\phi_n
  \right)^2
  \approx
  \frac{1}{N}
  \sum_{n=1}^N
  \left(
    y_n-\mu_n
  \right)^2
  +
  \frac{1}{N}
  \sum_{n=1}^N
  \left(
    \mu_n-\phi_n
  \right)^2.
\end{equation*}

Proposition \ref{prop:quadratic-asymptotic}
is obtained by applying Proposition \ref{prop:quadratic-RKHS}
to large (``universal'') RKHS.
The details
\ifnotCONF
  will be given in \S\ref{sec:proofs},
\fi
and here we will only demonstrate this idea
with a simple but non-trivial example.
Let $k$ and $m$ be positive integer constants such that $m>k/2$.
A prediction strategy $F$ will be included in $\FFF$
if its predictions $\phi_n$ satisfy
\begin{equation*}
  \phi_n
  =
  \begin{cases}
    0 & \text{if $n\le k$}\\
    f(y_{n-k},\ldots,y_{n-1}) & \text{otherwise},
  \end{cases}
\end{equation*}
where $f$ is a function from the Sobolev space $W^{m,2}([-Y,Y]^k)$
(see, e.g., \cite{\AdamsFournier} for the definition and properties of Sobolev spaces);
$\left\|F\right\|_{\FFF}$ is defined to be the Sobolev norm of $f$.
Every continuous function of $(y_{n-k},\ldots,y_{n-1})$
can be arbitrarily well approximated by functions in $W^{m,2}([-Y,Y]^k)$,
and so $\FFF$ is a suitable class of prediction strategies
if we believe that neither $x_1,\ldots,x_{n}$ nor $y_1,\ldots,y_{n-k-1}$
are useful in predicting $y_n$.

\subsection*{Very large benchmark classes}

Some interesting benchmark classes of prediction strategies
are too large to equip with the structure of RKHS
\cite{\DFVI}.
However, an analogue of Proposition \ref{prop:quadratic-RKHS}
can also be proved for some Banach spaces $\FFF$ of prediction strategies
(with the pointwise operations of addition and multiplication by scalar)
for which the constant $\ccc_{\FFF}$ defined by (\ref{eq:embedding-constant})
is finite.
The \emph{modulus of convexity} of a Banach space $U$
is defined as the function
\begin{equation*}
  \delta_{U}(\epsilon)
  :=
  \inf_{\substack{u,v\in S_{U}\\\left\|u-v\right\|_{U}=\epsilon}}
  \left(
    1
    -
    \left\|
      \frac{u+v}{2}
    \right\|_{U}
  \right),
  \quad
  \epsilon\in(0,2],
\end{equation*}
where
$
  S_{U}
  :=
  \left\{
    u\in U
    \st
    \left\|
      u
    \right\|_{U}
    =
    1
  \right\}
$
is the unit sphere in $U$.

The existence of leading strategies
(in a somewhat weaker sense than in Proposition \ref{prop:quadratic-RKHS})
is asserted in the following result.
\begin{proposition}\label{prop:quadratic-PBFS}
  Let $\mathbf{Y}=[-Y,Y]$, $\mathbf{P}=\bbbr$,
  and $\FFF$ be a Banach space of prediction strategies
  having a finite embedding constant $\ccc_{\FFF}$
  (see (\ref{eq:embedding-constant}))
  and satisfying
  \begin{equation*}
    \forall\epsilon\in(0,2]:
    \delta_{\FFF}(\epsilon)
    \ge
    (\epsilon/2)^p/p
  \end{equation*}
  for some $p\in[2,\infty)$.
  There exists a strategy for Predictor that guarantees
  \begin{multline}\label{eq:quadratic-PBFS}
    \left|
      \sum_{n=1}^N
      \left(
        y_n-\mu_n
      \right)^2
      +
      \sum_{n=1}^N
      \left(
        \mu_n-\phi_n
      \right)^2
      -
      \sum_{n=1}^N
      \left(
        y_n-\phi_n
      \right)^2
    \right|\\
    \le
    40Y
    \sqrt{\ccc_{\FFF}^2+1}
    \left(
      \left\|
        F
      \right\|_{\FFF}
      +
      Y
    \right)
    N^{1-1/p},
    \qquad
    \forall N\in\{1,2,\ldots\}
    \enspace
    \forall F\in\FFF,
  \end{multline}
  where $\phi_n$ are $F$'s predictions.
\end{proposition}
The example of a benchmark class of prediction strategies
given after Proposition \ref{prop:quadratic-RKHS}
but with $f$ ranging over the Sobolev space $W^{s,p}([-Y,Y]^k)$,
$s>k/p$,
is covered by this proposition.
The parameter $s$ describes the ``degree of regularity''
of the elements of $W^{s,p}$,
and taking sufficiently large $p$
we can reach arbitrarily irregular functions in the Sobolev hierarchy.

\section{Predictions evaluated by Bregman divergences}
\label{sec:bregman}


A \emph{predictable process} is a function $F$
mapping the situation space $\mathbf{S}$ to $\bbbr$,
$F:\mathbf{S}\to\bbbr$.
Notice that for any function $\psi:\mathbf{P}\to\bbbr$
and any prediction strategy $F$
the composition $\psi(F)$
(mapping each situation $s$ to $\psi(F(s))$)
is a predictable process;
such compositions will be used in Theorems \ref{thm:bregman}--\ref{thm:log} below.
A Hilbert space $\FFF$ of predictable processes
(with the usual pointwise operations)
is called an \emph{RKHS with finite embedding constant}
if (\ref{eq:embedding-constant}) is finite.

The notion of Bregman divergence was introduced in \cite{bregman:1967},
and is now widely used in competitive on-line prediction
(see, e.g.,
\cite{helmbold/etal:1999,azoury/warmuth:2001,%
herbster/warmuth:2001,kivinen/warmuth:2001,cesabianchi/lugosi:2006}).
Suppose $\mathbf{Y}=\mathbf{P}\subseteq\bbbr$
(although it would be interesting to extend Theorem \ref{thm:bregman}
to the case where $\bbbr$ is replaced by any Euclidean, or even Hilbert, space).
Let $\Psi$ and $\Psi'$ be two real-valued functions defined on $\mathbf{Y}$.
The expression
\begin{equation}\label{eq:def-bregman}
  d_{\Psi,\Psi'}(y,z)
  :=
  \Psi(y)
  -
  \Psi(z)
  -
  \Psi'(z)(y-z),
  \quad
  y,z\in\mathbf{Y},
\end{equation}
is said to be the corresponding \emph{Bregman divergence}
if $d_{\Psi,\Psi'}(y,z)>0$ whenever $y\ne z$.
(Bregman divergence is usually defined
for $y$ and $z$ ranging over a Euclidean space.)
In all our examples $\Psi$ will be a strictly convex continuously differentiable function
and $\Psi'$ its derivative,
in which case we abbreviate $d_{\Psi,\Psi'}$ to $d_{\Psi}$.

We will be using the standard notation
\begin{equation*}
  \left\|
    f
  \right\|_{C(A)}
  :=
  \sup_{y\in A}
  \left|
    f(y)
  \right|,
\end{equation*}
where $A$ is a subset of the domain of $f$.
\begin{theorem}\label{thm:bregman}
  Suppose $\mathbf{Y}=\mathbf{P}$ is a bounded subset of $\bbbr$.
  Let $\FFF$ be an RKHS of predictable processes
  with finite embedding constant $\ccc_{\FFF}$
  and $\Psi,\Psi'$ be real-valued functions on $\mathbf{Y}=\mathbf{P}$.
  There exists a strategy for Predictor that guarantees,
  for all prediction strategies $F$ and $N=1,2,\ldots$,
  \begin{multline}\label{eq:bregman}
    \left|
      \sum_{n=1}^N
      d_{\Psi,\Psi'}
      \left(
        y_n,\mu_n
      \right)
      +
      \sum_{n=1}^N
      d_{\Psi,\Psi'}
      \left(
        \mu_n,\phi_n
      \right)
      -
      \sum_{n=1}^N
      d_{\Psi,\Psi'}
      \left(
        y_n,\phi_n
      \right)
    \right|\\
    \le
    \diam(\mathbf{Y})
    \sqrt{\ccc_{\FFF}^2+1}
    \left(
      \left\|
        \Psi'(F)
      \right\|_{\FFF}
      +
      \left\|
        \Psi'
      \right\|_{C(\mathbf{Y})}
    \right)
    \sqrt{N},
  \end{multline}
  where $\phi_n$ are $F$'s predictions.
\end{theorem}
The expression
$
  \left\|
    \Psi'(F)
  \right\|_{\FFF}
$
in (\ref{eq:bregman}) is interpreted as $\infty$
when $\Psi'(F)\notin\FFF$;
in this case (\ref{eq:bregman}) holds vacuously.
Similar conventions will be made in all following statements.

Two of the most important Bregman divergences
are obtained from the convex functions
$\Psi(y):=y^2$ and $\Psi(y):=y\ln y+(1-y)\ln(1-y)$
(negative entropy,
defined for $y\in(0,1)$);
they are the quadratic loss function
\begin{equation}\label{eq:squared-difference}
  d_{\Psi}(y,z)
  =
  (y-z)^2
\end{equation}
and the relative entropy
(also known as the Kullback--Leibler divergence)
\begin{equation}\label{eq:KL}
  d_{\Psi}(y,z)
  =
  D(y\Parallel z)
  :=
  y\ln\frac{y}{z}
  +
  (1-y)\ln\frac{1-y}{1-z},
\end{equation}
respectively.
If we apply Theorem \ref{thm:bregman} to them,
(\ref{eq:squared-difference}) leads (assuming $\mathbf{Y}=[-Y,Y]$)
to a weaker version of Proposition \ref{prop:quadratic-RKHS},
with the right-hand side of (\ref{eq:bregman}) twice as large
as that of (\ref{eq:quadratic-RKHS}),
and (\ref{eq:KL}) leads to the following corollary.
\begin{corollary}\label{cor:KL}
  Let $\epsilon\in(0,1/2)$,
  $\mathbf{Y}=\mathbf{P}=[\epsilon,1-\epsilon]$,
  and the loss function be
  \begin{equation*}
    \lambda(y,\mu)
    =
    D(y\Parallel\mu)
  \end{equation*}
  (defined in (\ref{eq:KL})).
  Let $\FFF$ be an RKHS of predictable processes
  with finite embedding constant $\ccc_{\FFF}$.
  There exists a strategy for Predictor that guarantees,
  for all prediction strategies $F$,
  \begin{multline*}
    \left|
      \sum_{n=1}^N
      \lambda
      \left(
        y_n,\mu_n
      \right)
      +
      \sum_{n=1}^N
      \lambda
      \left(
        \mu_n,\phi_n
      \right)
      -
      \sum_{n=1}^N
      \lambda
      \left(
        y_n,\phi_n
      \right)
    \right|\\
    \le
    \sqrt{\ccc_{\FFF}^2+1}
    \left(
      \left\|
        \ln\frac{F}{1-F}
      \right\|_{\FFF}
      +
      \ln\frac{1-\epsilon}{\epsilon}
    \right)
    \sqrt{N},
    \qquad
    \forall N\in\{1,2,\ldots\},
  \end{multline*}
  where $\phi_n$ are $F$'s predictions.
\end{corollary}
The log likelihood ratio $\ln\frac{F}{1-F}$ appears
because $\Psi'(y)=\ln\frac{y}{1-y}$ in this case.

Analogously to Proposition \ref{prop:quadratic-RKHS},
Theorem \ref{thm:bregman}
(as well as Theorems \ref{thm:proper}--\ref{thm:log} in the next section)
can be easily generalized to Banach spaces of predictable processes.
One can also state asymptotic versions of Theorems \ref{thm:bregman}--\ref{thm:log}
similar to Proposition \ref{prop:quadratic-asymptotic};
and the continuous limited-memory strategies of Proposition \ref{prop:quadratic-asymptotic}
could be replaced by the equally interesting classes
of continuous stationary strategies (as in \cite{\DFVIII})
or Markov strategies (possibly discontinuous, as in \cite{\DFIX}).
We will have to refrain from pursuing these developments in this paper.

\section{Predictions evaluated by strictly proper scoring rules}
\label{sec:proper}

In this section we consider the case where $\mathbf{Y}=\{0,1\}$
and $\mathbf{P}\subseteq[0,1]$.
Every loss function $\lambda:\mathbf{Y}\times\mathbf{P}\to\bbbr$
will be extended to the domain $[0,1]\times\mathbf{P}$ by the formula
\begin{equation*}
  \lambda(p,\mu)
  :=
  p\lambda(1,\mu)+(1-p)\lambda(0,\mu);
\end{equation*}
intuitively,
$\lambda(p,\mu)$ is the expected loss of the prediction $\mu$
when the probability of $y=1$ is $p$.
Let us say that a loss function $\lambda$ is a \emph{strictly proper scoring rule}
if
\begin{equation*}
  \forall p,\mu\in\mathbf{P}:
  p\ne\mu
  \Longrightarrow
  \lambda(p,p)
  <
  \lambda(p,\mu)
\end{equation*}
(it is optimal to give the prediction equal to the true probability of $y=1$
when the latter is known and belongs to $\mathbf{P}$).
In this case the function
\begin{equation*}
  d_{\lambda}(\mu,\phi)
  :=
  \lambda(\mu,\phi)
  -
  \lambda(\mu,\mu)
\end{equation*}
can serve as a measure of difference between predictions $\mu$ and $\phi$:
it is non-negative and is zero only when $\mu=\phi$.
(Cf.\ \cite{dawid:1994}, \S4.)


The \emph{exposure} of a loss function $\lambda$ is defined as
\begin{equation*}
  \Exp_{\lambda}(\mu)
  :=
  \lambda(1,\mu)
  -
  \lambda(0,\mu),
  \quad
  \mu\in\mathbf{P}.
\end{equation*}

\begin{theorem}\label{thm:proper}
  Let $\mathbf{Y}=\{0,1\}$, $\mathbf{P}\subseteq[0,1]$,
  $\lambda$ be a strictly proper scoring rule,
  and $\FFF$ be an RKHS of predictable processes
  with finite embedding constant $\ccc_{\FFF}$.
  There exists a strategy for Predictor that guarantees,
  for all prediction strategies $F$ and all $N=1,2,\ldots$,
  \begin{multline}\label{eq:proper}
    \left|
      \sum_{n=1}^N
      \lambda
      \left(
        y_n,\mu_n
      \right)
      +
      \sum_{n=1}^N
      d_{\lambda}
      \left(
        \mu_n,\phi_n
      \right)
      -
      \sum_{n=1}^N
      \lambda
      \left(
        y_n,\phi_n
      \right)
    \right|\\
    \le
    \frac{\sqrt{\ccc_{\FFF}^2+1}}{2}
    \left(
      \left\|
        \Exp_{\lambda}(F)
      \right\|_{\FFF}
      +
      \left\|
        \Exp_{\lambda}
      \right\|_{C(\mathbf{P})}
    \right)
    \sqrt{N},
  \end{multline}
  where $\phi_n$ are $F$'s predictions.
\end{theorem}
Two popular strictly proper scoring rules
are the quadratic loss function $\lambda(y,\mu):=(y-\mu)^2$
and the log loss function
\begin{equation*}
  \lambda(y,\mu)
  :=
  \begin{cases}
    -\ln\mu & \text{if $y=1$}\\
    -\ln(1-\mu) & \text{if $y=0$}.
  \end{cases}
\end{equation*}
Applied to the quadratic loss function,
Theorem \ref{thm:proper} becomes essentially
a special case of Proposition \ref{prop:quadratic-RKHS}.
For the log loss function we have
$d_{\lambda}(\mu,\phi)=D(\mu\Parallel\phi)$,
and so we obtain the following corollary.
\begin{corollary}\label{cor:log}
  Let $\epsilon\in(0,1/2)$, $\mathbf{Y}=\{0,1\}$, $\mathbf{P}=[\epsilon,1-\epsilon]$,
  $\lambda$ be the log loss function,
  and $\FFF$ be an RKHS of predictable processes
  with finite embedding constant $\ccc_{\FFF}$.
  There exists a strategy for Predictor that guarantees,
  for all prediction strategies $F$,
  \begin{multline*}
    \left|
      \sum_{n=1}^N
      \lambda
      \left(
        y_n,\mu_n
      \right)
      +
      \sum_{n=1}^N
      D
      \left(
        \mu_n\Parallel\phi_n
      \right)
      -
      \sum_{n=1}^N
      \lambda
      \left(
        y_n,\phi_n
      \right)
    \right|\\
    \le
    \frac{\sqrt{\ccc_{\FFF}^2+1}}{2}
    \left(
      \left\|
        \ln\frac{F}{1-F}
      \right\|_{\FFF}
      +
      \ln\frac{1-\epsilon}{\epsilon}
    \right)
    \sqrt{N},
    \qquad
    \forall N\in\{1,2,\ldots\},
  \end{multline*}
  where $\phi_n$ are $F$'s predictions.
\end{corollary}

A weaker version (with the bound twice as large)
of Corollary \ref{cor:log} would be a special case of Corollary \ref{cor:KL}
were it not for the restriction of the observation space $\mathbf{Y}$
to $[\epsilon,1-\epsilon]$ in the latter.
Using methods of \cite{\DFIV},
it is even possible to get rid of the restriction $\mathbf{P}=[\epsilon,1-\epsilon]$
in Corollary \ref{cor:log}.
Since the log loss function plays a fundamental role
in information theory
(the cumulative loss corresponds to the code length),
we state this result as our next theorem.
\begin{theorem}\label{thm:log}
  Let $\mathbf{Y}=\{0,1\}$, $\mathbf{P}=(0,1)$,
  $\lambda$ be the log loss function,
  and $\FFF$ be an RKHS of predictable processes
  with finite embedding constant $\ccc_{\FFF}$.
  There exists a strategy for Predictor that guarantees,
  for all prediction strategies $F$,
  \begin{multline*}
    \left|
      \sum_{n=1}^N
      \lambda
      \left(
        y_n,\mu_n
      \right)
      +
      \sum_{n=1}^N
      D
      \left(
        \mu_n\Parallel\phi_n
      \right)
      -
      \sum_{n=1}^N
      \lambda
      \left(
        y_n,\phi_n
      \right)
    \right|\\
    \le
    \frac{\sqrt{\ccc_{\FFF}^2+1.8}}{2}
    \left(
      \left\|
        \ln\frac{F}{1-F}
      \right\|_{\FFF}
      +
      1
    \right)
    \sqrt{N},
    \qquad
    \forall N\in\{1,2,\ldots\},
  \end{multline*}
  where $\phi_n$ are $F$'s predictions.
\end{theorem}

\section{Stochastic Reality and Jeffreys's law}
\label{sec:stochastic}

In this section we revert to the quadratic regression framework of \S\ref{sec:regression}
and assume $\mathbf{Y}=\mathbf{P}=[-Y,Y]$,
$\lambda(y,\mu)=(y-\mu)^2$.
(It will be clear that similar results hold for Bregman divergences
and strictly proper scoring rules,
but we stick to the simplest case
since our main goal in this section
is to discuss the related literature.)
\begin{proposition}\label{prop:quadratic-true}
  Suppose $\mathbf{Y}=\mathbf{P}=[-Y,Y]$.
  Let $F$ be a prediction strategy
  and $y_n\in[-Y,Y]$ be generated as
  $
    y_n
    :=
    F(s_n)
    +
    \xi_n
  $
  (remember that $s_n$ are defined by (\ref{eq:s-n})),
  where the noise random variables $\xi_n$ have expected value zero given $s_n$.
  For any other prediction strategy $G$,
  any $N\in\{1,2,\ldots\}$, and any $\delta\in(0,1)$,
  \begin{equation}\label{eq:quadratic-true}
    \left|
      \sum_{n=1}^N
      \left(
        y_n-\phi_n
      \right)^2
      +
      \sum_{n=1}^N
      \left(
        \phi_n-\mu_n
      \right)^2
      -
      \sum_{n=1}^N
      \left(
        y_n-\mu_n
      \right)^2
    \right|
    \le
    4Y^2
    \sqrt{2\ln\frac{2}{\delta}}
    \sqrt{N}
  \end{equation}
  with probability at least $1-\delta$,
  where $\phi_n$ are $F$'s predictions and $\mu_n$ are $G$'s predictions.
\end{proposition}
Combining Proposition \ref{prop:quadratic-true} with Proposition \ref{prop:quadratic-RKHS}
we obtain the following corollary.
\begin{corollary}\label{cor:jeffreys}
  Suppose $\mathbf{Y}=\mathbf{P}=[-Y,Y]$.
  Let $\FFF$ be an RKHS of prediction strategies
  with finite embedding constant $\ccc_{\FFF}$,
  $G$ be a prediction strategy whose predictions $\mu_n$
  are guaranteed to satisfy (\ref{eq:quadratic-RKHS})
  (a ``leading prediction strategy''),
  $F$ be a prediction strategy in $\FFF$,
  and $y_n\in[-Y,Y]$ be generated as
  $
    y_n
    :=
    F(s_n)
    +
    \xi_n,
  $
  where the noise random variables $\xi_n$ have expected value zero given $s_n$.
  For any $N\in\{1,2,\ldots\}$ and any $\delta\in(0,1)$,
  the conjunction of
  \begin{multline}\label{eq:jeffreys1}
    \left|
      \sum_{n=1}^N
      \left(
        y_n-\mu_n
      \right)^2
      -
      \sum_{n=1}^N
      \left(
        y_n-\phi_n
      \right)^2
    \right|\\
    \le
    Y
    \sqrt{\ccc_{\FFF}^2+1}
    \left(
      \left\|
        F
      \right\|_{\FFF}
      +
      Y
    \right)
    \sqrt{N}
    +
    2Y^2
    \sqrt{2\ln\frac{2}{\delta}}
    \sqrt{N}
  \end{multline}
  and
  \begin{equation}\label{eq:jeffreys2}
    \sum_{n=1}^N
    \left(
      \phi_n-\mu_n
    \right)^2
    \le
    Y
    \sqrt{\ccc_{\FFF}^2+1}
    \left(
      \left\|
        F
      \right\|_{\FFF}
      +
      Y
    \right)
    \sqrt{N}
    +
    2Y^2
    \sqrt{2\ln\frac{2}{\delta}}
    \sqrt{N}
  \end{equation}
  holds with probability at least $1-\delta$,
  where $\phi_n$ are $F$'s predictions and $\mu_n$ are $G$'s predictions.
\end{corollary}
We can see that if the ``true'' (in the sense of outputting the true expectations)
strategy $F$ belongs to the RKHS $\FFF$
and $\left\|F\right\|_{\FFF}$ is not too large,
not only the loss of the leading strategy will be close to that of the true strategy,
but their predictions will be close as well.

\subsection*{Jeffreys's law}

In the rest of this section we will explain the connection of this paper
with the phenomenon widely studied in probability theory
and the algorithmic theory of randomness
and dubbed ``Jeffreys's law'' by Dawid \cite{dawid:1984,dawid:2004}.
The general statement of ``Jeffreys's law''
is that two successful prediction strategies produce similar predictions
(cf.\ \cite{dawid:1984}, \S5.2).
To better understand this informal statement,
we first discuss two notions of success for prediction strategies.

As argued in \cite{\DFVII},
there are (at least) two very different kinds of predictions,
which we will call ``S-predictions'' and ``D-predictions''.
Both S-predictions and D-predictions are elements of $[-Y,Y]$
(in our current context),
and the prefixes ``S-'' and ``D-'' refer to the way
in which we want to evaluate their quality.
S-predictions are Statements about Reality's behaviour,
and they are successful if they withstand attempts to falsify them;
standard means of falsification are statistical tests
(see, e.g., \cite{cox/hinkley:1974}, Chapter 3)
and gambling strategies
(\cite{ville:1939};
for a more recent exposition, see \cite{shafer/vovk:2001}).
D-predictions do not claim to be falsifiable statements about Reality;
they are Decisions deemed successful
if they lead to a good cumulative loss.

As an example,
let us consider the predictions $\phi_n$ and $\mu_n$
in Proposition \ref{prop:quadratic-true}.
The former are S-predictions;
they can be rejected
if (\ref{eq:quadratic-true}) fails to happen for a small $\delta$
(the complement of (\ref{eq:quadratic-true})
can be used as the critical region of a statistical test).
The latter are D-predictions:
we are only interested in their cumulative loss.
If $\phi_n$ are successful
((\ref{eq:quadratic-true}) holds for a moderately small $\delta$)
and $\mu_n$ are successful
(in the sense of their cumulative loss being close
to the cumulative loss of the successful S-predictions $\phi_n$;
this is the best that can be achieved
as, by (\ref{eq:quadratic-true}),
the latter cannot be much larger than the former),
they will be close to each other,
in the sense $\sum_{n=1}^{N}(\phi_n-\mu_n)^2\ll N$.
We can see that Proposition \ref{prop:quadratic-true}
implies a ``mixed'' version of Jeffreys's law,
asserting the proximity of S-predictions and D-predictions.

Similarly, Corollary \ref{cor:jeffreys}
is also a mixed version of Jeffreys's law:
it asserts the proximity of the S-predictions $\phi_n$
(which are part of our falsifiable model $y_n=\phi_n+\xi_n$)
and the D-predictions $\mu_n$
(successful in the sense of leading to a good cumulative loss;
cf.\ (\ref{eq:quadratic-RKHS})).

Proposition \ref{prop:quadratic-RKHS} immediately implies
two ``pure'' versions of Jeffreys's laws for D-predictions:
\begin{itemize}
\item
  if a prediction strategy $F$ with $\left\|F\right\|_{\FFF}$ not too large
  performs well, in the sense that its loss is close to the leading strategy's loss,
  $F$'s predictions will be similar to the leading strategy's predictions;
  more precisely,
  \begin{multline*}
    \sum_{n=1}^N
    \left(
      \phi_n-\mu_n
    \right)^2
    \le
    \sum_{n=1}^N
    \left(
      y_n-\phi_n
    \right)^2
    -
    \sum_{n=1}^N
    \left(
      y_n-\mu_n
    \right)^2\\
    +
    2Y
    \sqrt{\ccc_{\FFF}^2+1}
    \left(
      \left\|
        F
      \right\|_{\FFF}
      +
      Y
    \right)
    \sqrt{N};
  \end{multline*}
\item
  therefore,
  if two prediction strategies $F_1$ and $F_2$
  with $\left\|F_1\right\|_{\FFF}$ and $\left\|F_2\right\|_{\FFF}$ not too large
  perform well, in the sense that their loss is close to the leading strategy's loss,
  their predictions will be similar.
\end{itemize}
It is interesting that the leading strategy can be replaced by a master strategy
for the second version:
if $F_1$ and $F_2$ gave very different predictions
and both performed almost as well as the master strategy,
the mixed strategy $(F_1+F_2)/2$ would beat the master strategy;
this immediately follows from
\begin{equation*}
  \left(
    \frac{\phi_1+\phi_2}{2}
    -
    y
  \right)^2
  =
  \frac{(\phi_1-y)^2+(\phi_2-y)^2}{2}
  -
  \left(
    \frac{\phi_1-\phi_2}{2}
  \right)^2,
\end{equation*}
where $\phi_1$ and $\phi_2$ are $F_1$'s and $F_2$'s predictions, respectively,
and $y$ is the observation.

The usual versions of Jeffreys's law are, however,
statements about S-predictions.
The quality of S-predictions is often evaluated
using universal statistical tests
(as formalized by Martin-L\"of \cite{martin-lof:1966})
or universal gambling strategies
(Levin \cite{levin:1973random}, Schnorr \cite{schnorr:1971book}).
For example,
Theorem 7.1 of \cite{dawid:1985}
and Theorem 3 of \cite{vovk:1987criterion}
state that if two computable S-prediction strategies
are both successful,
their predictions will asymptotically agree.
Earlier, somewhat less intuitive, statements of Jeffreys's law
were given in terms of absolute continuity of probability measures:
see, e.g.,
\cite{blackwell/dubins:1962} and \cite{\KabanovEtAl}.
Solomonoff \cite{solomonoff:1978} proved a version of Jeffreys's law that holds
``on average''
(rather than for individual sequences).

This paper is, to my knowledge, the first to state a version of Jeffreys's law
for D-predictions
(although a step in this direction was made
in Theorem~8 of \cite{\VovkTCS}).

\section{Proofs}
\label{sec:proofs}

\ifnotCONF
  In this section we prove,
  or give proof sketches of,
  Propositions \ref{prop:quadratic-asymptotic}--\ref{prop:quadratic-true}
  and Theorems \ref{thm:bregman}--\ref{thm:log}.
  Proposition \ref{prop:quadratic-RKHS} is a special case of Theorem \ref{thm:bregman},
  but its proof is more intuitive and we give it separately
  (proving Proposition \ref{prop:quadratic-PBFS} along the way).
\fi

\subsection*{Proof of Propositions \ref{prop:quadratic-RKHS} and \ref{prop:quadratic-PBFS}}

Noticing that
\begin{multline}\label{eq:basic-quadratic}
  \left|
    \sum_{n=1}^N
    \left(
      y_n-\mu_n
    \right)^2
    +
    \sum_{n=1}^N
    \left(
      \mu_n-\phi_n
    \right)^2
    -
    \sum_{n=1}^N
    \left(
      y_n-\phi_n
    \right)^2
  \right|\\
  =
  2
  \left|
    \sum_{n=1}^N
    \left(
      \phi_n-\mu_n
    \right)
    \left(
      y_n-\mu_n
    \right)
  \right|\\
  \le
  2
  \left|
    \sum_{n=1}^N
    \mu_n
    \left(
      y_n-\mu_n
    \right)
  \right|
  +
  2
  \left|
    \sum_{n=1}^N
    \phi_n
    \left(
      y_n-\mu_n
    \right)
  \right|,
\end{multline}
we can use the results of \cite{\DFV}, \S6,
asserting the existence of a prediction strategy
producing predictions $\mu_n\in[-Y,Y]$ that satisfy
\begin{equation}\label{eq:calibration1}
  \left|
    \sum_{n=1}^N
    \mu_n
    \left(
      y_n-\mu_n
    \right)
  \right|
  \le
  Y^2
  \sqrt{\ccc_{\FFF}^2+1}
  \sqrt{N}
\end{equation}
(see (24) in \cite{\DFV};
this a special case of good calibration)
and
\begin{equation}\label{eq:resolution1}
  \left|
    \sum_{n=1}^N
    \phi_n
    \left(
      y_n-\mu_n
    \right)
  \right|
  \le
  Y
  \sqrt{\ccc_{\FFF}^2+1}
  \left\|
    F
  \right\|_{\FFF}
  \sqrt{N}
\end{equation}
(see (25) in \cite{\DFV};
this a special case of good resolution).

Replacing (\ref{eq:calibration1}) and (\ref{eq:resolution1})
with the corresponding statements for Banach function spaces
(\cite{\DFVI}, (52) and (53))
we obtain the proof of Proposition \ref{prop:quadratic-PBFS}.

\begin{remark*}
  In \cite{\DFV} we considered only prediction strategies
  $F$ for which $F(s_n)$ depends on $s_n$ (see (\ref{eq:s-n})) via $x_n$;
  in the terminology of this paper these are (order 0) Markov strategies.
  It is easy to see that considering only Markov strategies
  does not lead to a loss of generality:
  if we redefine the object $x_n$ as
  $x_n:=s_n$,
  any prediction strategy
  will become a Markov prediction strategy.
\end{remark*}

\ifnotCONF
\subsection*{Proof of Proposition \ref{prop:quadratic-asymptotic}}

Proposition \ref{prop:quadratic-asymptotic} will follow
from the following lemma,
proved (without stating it explicitly)
in \cite{steinwart/etal:2006} (proof of Theorem 2).
\begin{lemma}[\cite{steinwart/etal:2006}]\label{eq:steinwart}
  Let $\GGG$ be a separable set in $C(Z)$.
  There exists an RKHS $\FFF$ on $Z$ with finite embedding constant
  such that $\FFF$ is dense in $\GGG$ in metric $C(Z)$.
\end{lemma}
\begin{proof}
  Let $F_1,F_2,\ldots$ be a dense (in metric $C(Z)$)
  sequence of elements of $\GGG$.
  Set
  \begin{equation*}
    \Phi_n
    :=
    \begin{cases}
      2^{-n}
      \left\|F_n\right\|_{C(Z)}^{-1}
      F_n
      & \text{if $F_n\ne0$}\\
      0
      & \text{otherwise},
    \end{cases}
  \end{equation*}
  $\Phi(z):=(\Phi_1(z),\Phi_2(z),\ldots)\in\ell_2$ for $z\in Z$,
  \begin{equation*}
    \KKK(z,z')
    :=
    \left\langle
      \Phi(z),
      \Phi(z')
    \right\rangle_{\ell_2},
    \qquad
    z,z'\in Z,
  \end{equation*}
  and let $\FFF$ be the unique RKHS
  with reproducing kernel $\KKK$
  (see the Moore--Aronszajn theorem
  in \cite{aronszajn:1943}, Theorem 2).
  It is clear that $\ccc^2_{\FFF}=\sup_z\KKK(z,z)$ is finite.
  By Lemma \ref{lem:kernel} below,
  each $F_n$ belongs to $\FFF$ since it can be represented as
  \begin{equation*}
    \left\langle
      2^{n}
      \left\|F_n\right\|_{C(Z)}
      e_n,
      \Phi(\cdot)
    \right\rangle_{\ell_2},
  \end{equation*}
  where $e_n\in\ell_2$ consists of all $0$s except a $1$ at the $n$th position.
  Therefore, $\FFF$ is dense in $\GGG$.
  \qedtext
\end{proof}
The following lemma was used in the proof.
\begin{lemma}\label{lem:kernel}
  Let $\Phi:Z\to H$,
  where $H$ is a Hilbert space.
  The RKHS corresponding to the reproducing kernel
  $\KKK(z,z'):=\langle\Phi(z),\Phi(z')\rangle_H$
  consists of all functions $\langle v,\Phi(\cdot)\rangle_H$, $v\in H$,
  with the inner product of $\langle v,\Phi(\cdot)\rangle_H$ and $\langle v',\Phi(\cdot)\rangle_H$
  equal to $\langle p(v),p(v')\rangle_H$,
  $p$ standing for the projection onto the span of $\Phi(Z)$.
\end{lemma}
\begin{proof}
  By the Moore--Aronszajn theorem
  (\cite{aronszajn:1943}, Theorem 2)
  there is a unique RKHS with reproducing kernel $\KKK$,
  so we only need to check that the function space $\FFF$
  defined in the statement of the lemma
  is an RKHS with $\KKK$ as reproducing kernel.

  First we need to check that the inner product is well defined.
  This follows from the obvious fact that the equality of the functions
  $\langle v,\Phi(\cdot)\rangle_H$ and $\langle v',\Phi(\cdot)\rangle_H$
  for $v,v'\in\Span(\Phi(Z))$
  implies $v=v'$.
  The continuity of each evaluation functional is also obvious.

  The representer of $z\in Z$ is
  $\KKK_z(\cdot):=\langle\Phi(z),\Phi(\cdot)\rangle_H$
  (in the sense that $\langle\KKK_z,f\rangle_{\FFF}=f(z)$ for each $f\in\FFF$)
  and so the reproducing kernel $\langle\KKK_z,\KKK_{z'}\rangle_{\FFF}$ of $\FFF$
  indeed coincides with $\KKK$.
  \qedtext
\end{proof}

Now we can easily deduce Proposition \ref{prop:quadratic-asymptotic}
from Proposition \ref{prop:quadratic-RKHS}.
The set of all continuous order $k$ Markov strategies
is a separable set in the Banach space $C(\mathbf{S})$
of continuous prediction strategies with the sup metric
(by \cite{engelking:1989}, Corollary 4.2.18).
Therefore, the set $\GGG$ of all continuous limited-memory strategies
is separable in $C(\mathbf{S})$.

Let $\FFF$ be the RKHS whose existence is asserted by Lemma \ref{eq:steinwart};
we will see that any strategy for Predictor satisfying (\ref{eq:quadratic-RKHS})
and $\mu_n\in[-Y,Y]$
will satisfy (\ref{eq:quadratic-asymptotic})
with $\phi_n$ output by a limited-memory strategy $F$.
Indeed, for any $\epsilon>0$ we can find $F^*\in\FFF$ that is $\epsilon$-close
in $C(\mathbf{S})$ to $G$.
If $\phi_n$ are $F$'s predictions
and $\phi_n^*$ are $F^*$'s predictions,
(\ref{eq:quadratic-RKHS}) implies that
\begin{multline*}
  \left|
    \frac1N
    \sum_{n=1}^N
    \left(
      y_n-\mu_n
    \right)^2
    +
    \frac1N
    \sum_{n=1}^N
    \left(
      \mu_n-\phi_n
    \right)^2
    -
    \frac1N
    \sum_{n=1}^N
    \left(
      y_n-\phi_n
    \right)^2
  \right|\\
  \le
  \left|
    \frac1N
    \sum_{n=1}^N
    \left(
      y_n-\mu_n
    \right)^2
    +
    \frac1N
    \sum_{n=1}^N
    \left(
      \mu_n-\phi^*_n
    \right)^2
    -
    \frac1N
    \sum_{n=1}^N
    \left(
      y_n-\phi^*_n
    \right)^2
  \right|
  +
  8(Y+\epsilon)\epsilon\\
  \le
  2Y
  \sqrt{\ccc_{\FFF}^2+1}
  \left(
    \left\|
      F
    \right\|_{\FFF}
    +
    Y
  \right)
  \frac{1}{\sqrt{N}}
  +
  8(Y+\epsilon)\epsilon
  \le
  10(Y+\epsilon)\epsilon
\end{multline*}
from some $N$ on.
Since $\epsilon$ can be taken arbitrarily small,
we have (\ref{eq:quadratic-asymptotic}).
\fi

\ifnotCONF
  \subsection*{Proof of Theorem \ref{thm:bregman}}
\fi

The proof is based on the \emph{generalized law of cosines}
\begin{equation}\label{eq:cosines}
  d_{\Psi,\Psi'}(y,\phi)
  =
  d_{\Psi,\Psi'}(\mu,\phi)
  +
  d_{\Psi,\Psi'}(y,\mu)
  -
  \left(
    \Psi'(\phi) - \Psi'(\mu)
  \right)
  \left(
    y-\mu
  \right)
\end{equation}
(which follows directly from the definition (\ref{eq:def-bregman})).
From (\ref{eq:cosines}) we deduce
\begin{multline}\label{eq:basic-bregman}
  \left|
    \sum_{n=1}^N
    d_{\Psi,\Psi'}
    \left(
      y_n,\mu_n
    \right)
    +
    \sum_{n=1}^N
    d_{\Psi,\Psi'}
    \left(
      \mu_n,\phi_n
    \right)
    -
    \sum_{n=1}^N
    d_{\Psi,\Psi'}
    \left(
      y_n,\phi_n
    \right)
  \right|\\
  =
  \left|
    \sum_{n=1}^N
    \left(
      \Psi'(\phi_n)
      -
      \Psi'(\mu_n)
    \right)
    \left(
      y_n-\mu_n
    \right)
  \right|\\
  \le
  \left|
    \sum_{n=1}^N
    \Psi'(\mu_n)
    \left(
      y_n-\mu_n
    \right)
  \right|
  +
  \left|
    \sum_{n=1}^N
    \Psi'(\phi_n)
    \left(
      y_n-\mu_n
    \right)
  \right|.
\end{multline}
\ifnotCONF
From Theorem 3 in \cite{\DFII} we can see
that there is a prediction strategy guaranteeing
\begin{equation}\label{eq:calibration2a}
  \left|
    \sum_{n=1}^N
    \Psi'(\mu_n)
    \left(
      y_n-\mu_n
    \right)
  \right|
  \le
  \diam(\mathbf{Y})
  \left\|
    \Psi'
  \right\|_{C(\mathbf{Y})}
  \sqrt{N}
\end{equation}
and from Theorem 4 in \cite{\DFII} we can see
that there is a prediction strategy guaranteeing
\begin{equation}\label{eq:resolution2a}
  \left|
    \sum_{n=1}^N
    \Psi'(\phi_n)
    \left(
      y_n-\mu_n
    \right)
  \right|
  \le
  \diam(\mathbf{Y})
  \ccc_{\FFF}
  \left\|
    \Psi'(F)
  \right\|_{\FFF}
  \sqrt{N}.
\end{equation}
We need, however, a single strategy guaranteeing
some versions of (\ref{eq:calibration2a}) and (\ref{eq:resolution2a}).
Such a strategy can be obtained by merging a strategy guaranteeing (\ref{eq:calibration2a})
and a strategy guaranteeing (\ref{eq:resolution2a})
(as in \cite{\DFV}, Corollaries 3 and 4).

Setting
\begin{equation}\label{eq:combined-mapping}
  \Phi(\mu,s)
  :=
  \left(
    \frac{\Psi'(\mu)}{\left\|\Psi'\right\|_{C(\mathbf{Y})}},
    \kkk_s
  \right)
  \in
  \bbbr\times\FFF,
  \quad
  \mu\in\mathbf{P},
  \enspace
  s\in\mathbf{S},
\end{equation}
so that $\ccc_{\Phi}\le\sqrt{\ccc_{\FFF}^2+1}$,
and letting $\mu_n$ be output by the K29 algorithm based on (\ref{eq:combined-mapping}),
we obtain
\begin{multline}\label{eq:calibration2b}
  \left|
    \sum_{n=1}^N
    \Psi'(\mu_n)
    \left(
      y_n-\mu_n
    \right)
  \right|
  \le
  \left\|
    \Psi'
  \right\|_{C(\mathbf{Y})}
  \left\|
    \sum_{n=1}^N
    \left(
      y_n-\mu_n
    \right)
    \Phi(\mu_n,s_n)
  \right\|_{\bbbr\times\FFF}\\
  \le
  \left\|
    \Psi'
  \right\|_{C(\mathbf{Y})}
  \diam(\mathbf{Y})
  \sqrt{\ccc_{\FFF}^2+1}
  \sqrt{N}
\end{multline}
from Theorem 3 of \cite{\DFII},
and we obtain
\begin{multline}\label{eq:resolution2b}
  \left|
    \sum_{n=1}^N
    \Psi'(\phi_n)
    \left(
      y_n-\mu_n
    \right)
  \right|
  =
  \left|
    \sum_{n=1}^N
    \left(
      y_n-\mu_n
    \right)
    \left\langle
      \kkk_{s_n},
      \Psi'(F)
    \right\rangle_{\FFF}
  \right|\\
  =
  \left|
    \left\langle
      \sum_{n=1}^N
      \left(
        y_n-\mu_n
      \right)
      \kkk_{s_n},
      \Psi'(F)
    \right\rangle_{\FFF}
  \right|
  \le
  \left\|
    \Psi'(F)
  \right\|_{\FFF}
  \left\|
    \sum_{n=1}^N
    \left(
      y_n-\mu_n
    \right)
    \kkk_{s_n}
  \right\|_{\FFF}\\
  \le
  \left\|
    \Psi'(F)
  \right\|_{\FFF}
  \left\|
    \sum_{n=1}^N
    \left(
      y_n-\mu_n
    \right)
    \Phi(\mu_n,s_n)
  \right\|_{\bbbr\times\FFF}\\
  \le
  \left\|
    \Psi'(F)
  \right\|_{\FFF}
  \diam(\mathbf{Y})
  \sqrt{\ccc_{\FFF}^2+1}
  \sqrt{N}
\end{multline}
from the proof of Theorem 4 and from Theorem 3 of \cite{\DFII}.

Combining (\ref{eq:basic-bregman})
with (\ref{eq:calibration2b}) and (\ref{eq:resolution2b})
we can see that (\ref{eq:combined-mapping})
produces a strategy guaranteeing (\ref{eq:bregman}).

\begin{remark*}
  As we mentioned earlier,
  the leading constant in the bound of Theorem \ref{thm:bregman}
  (and its corollary)
  is worse than those in other results in this paper,
  in the intersection of their domains of application.
  The explanation is that Theorem \ref{thm:bregman} is based on the K29 algorithm,
  whereas all other results are based on the more sophisticated
  ``K29${}^*$ algorithm''.
\end{remark*}
\fi

\subsection*{Proof sketch of Theorem \ref{thm:proper}}

The proof is similar to that of Theorem \ref{thm:bregman},
with the role of the generalized law of cosines (\ref{eq:cosines})
played by the equation
\begin{equation}\label{eq:to-solve}
  \lambda(y,\phi)
  =
  a
  +
  \lambda(y,\mu)
  +
  b(y-\mu)
\end{equation}
for some $a=a(\mu,\phi)$ and $b=b(\mu,\phi)$.
Since $y$ can take only two possible values,
suitable $a$ and $b$ are easy to find:
it suffices to solve the linear system
\begin{equation*}
  \begin{cases}
    \lambda(1,\phi)
    =
    a + \lambda(1,\mu) + b(1-\mu)\\

    \lambda(0,\phi)
    =
    a + \lambda(0,\mu) + b(-\mu).
  \end{cases}
\end{equation*}
Subtracting these equations we obtain
$
  b
  =
  \Exp(\phi) - \Exp(\mu)
$
(abbreviating $\Exp_{\lambda}$ to $\Exp$),
which in turn gives
$
  a
  =
  d_{\lambda}(\mu,\phi)$.
Therefore,
(\ref{eq:to-solve}) gives
\begin{multline}\label{eq:bound}
  \left|
    \sum_{n=1}^N
    \lambda
    \left(
      y_n,\mu_n
    \right)
    +
    \sum_{n=1}^N
    d_{\lambda}
    \left(
      \mu_n,\phi_n
    \right)
    -
    \sum_{n=1}^N
    \lambda
    \left(
      y_n,\phi_n
    \right)
  \right|\\
  =
  \left|
    \sum_{n=1}^N
    \left(
      \Exp(\phi_n)
      -
      \Exp(\mu_n)
    \right)
    \left(
      y_n-\mu_n
    \right)
  \right|\\
  \le
  \left|
    \sum_{n=1}^N
    \Exp(\mu_n)
    \left(
      y_n-\mu_n
    \right)
  \right|
  +
  \left|
    \sum_{n=1}^N
    \Exp(\phi_n)
    \left(
      y_n-\mu_n
    \right)
  \right|.
\end{multline}
\ifnotCONF
There are prediction strategies that guarantee
\begin{equation}\label{eq:calibration3}
  \left|
    \sum_{n=1}^N
    \Exp(\mu_n)
    \left(
      y_n-\mu_n
    \right)
  \right|
  \le
  \frac12
  \left\|
    \Exp
  \right\|_{C(\mathbf{P})}
  \sqrt{N}
\end{equation}
(cf.\ \cite{\DFIV}, Theorem 2)
and there are prediction strategies that guarantee
\begin{equation}\label{eq:resolution3}
  \left|
    \sum_{n=1}^N
    \Exp(F(s_n))
    \left(
      y_n-\mu_n
    \right)
  \right|
  \le
  \frac{\ccc_{\FFF}}{2}
  \left\|
    \Exp(F)
  \right\|_{\FFF}
  \sqrt{N}
\end{equation}
(cf.\ \cite{\DFIV}, Theorem 3);
merging such strategies as in \cite{\DFV}, Corollaries 3 and 4,
we can easily obtain (\ref{eq:proper})
from (\ref{eq:bound}), (\ref{eq:calibration3}), and (\ref{eq:resolution3}).

\subsection*{Proof sketch of Theorem \ref{thm:log}}

It is shown in \cite{\DFIV} that there is a prediction strategy guaranteeing
\begin{equation}\label{eq:calibration4}
  \left|
    \sum_{n=1}^N
    \Exp(\mu_n)
    (y_n-\mu_n)
  \right|
  \le
  \sqrt
  {
    \sum_{n=1}^N
    \mu_n(1-\mu_n)
    \left(
      \Exp^2(\mu_n)
      +
      \KKK(s_n,s_n)
    \right)
  }
\end{equation}
and
\begin{multline}\label{eq:resolution4}
  \left|
    \sum_{n=1}^N
    \Exp(F(s_n))
    (y_n-\mu_n)
  \right|\\
  \le
  \left\|
    \Exp(F)
  \right\|_{\FFF}
  \sqrt
  {
    \sum_{n=1}^N
    \mu_n(1-\mu_n)
    \left(
      \Exp^2(\mu_n)
      +
      \KKK(s_n,s_n)
    \right)
  }
\end{multline}
(see (21), (22), and the subsection ``Proof: Part II''
in \cite{\DFIV}, the technical report),
where $\KKK$ is the reproducing kernel of $\FFF$.
Comparing (\ref{eq:calibration4}) and (\ref{eq:resolution4}) with (\ref{eq:bound}),
we can see that Theorem \ref{thm:log} will follow from
\begin{equation*}
  \sqrt
  {
    \sum_{n=1}^N
    \mu_n(1-\mu_n)
    \left(
      \Exp^2(\mu_n)
      +
      \KKK(s_n,s_n)
    \right)
  }
  \le
  \frac{\sqrt{\ccc_{\FFF}^2+1.8}}{2}
  \sqrt{N},
\end{equation*}
which in turn will follow from
\begin{equation*}
  \mu(1-\mu)
  \left(
    \ln^2\frac{\mu}{1-\mu}
    +
    \ccc_{\FFF}^2
  \right)
  \le
  \frac{\ccc_{\FFF}^2+1.8}{4}.
\end{equation*}
It remains to notice that $\mu(1-\mu)\le1/4$ and to calculate
\begin{equation*}
  \sup_{\mu}
  \left(
    4\mu(1-\mu)
    \ln^2\frac{\mu}{1-\mu}
  \right)
  \approx
  1.76
  \le
  1.8.
\end{equation*}

\subsection*{Proof of Proposition \ref{prop:quadratic-true}}

This proposition immediately follows from the equality in (\ref{eq:basic-quadratic})
and Hoeffding's inequality
(see, e.g., \cite{devroye/etal:1996}, p.~135).
\fi

\section{Conclusion}
\label{sec:conclusion}

The existence of master strategies
(strategies whose loss is less than or close to the loss of any strategy with not too large a norm)
can be shown for a very wide class of loss functions.
On the contrary,
leading strategies appear to exist for a rather narrow class of loss functions.
It would be very interesting to delineate the class of loss functions
for which a leading strategy does exist.
In particular,
does this class contain any loss functions except Bregman divergences
and strictly proper scoring rules?

Even if a leading strategy does not exist,
one might look for a strategy $G$ such that the loss of any strategy $F$
whose norm is not too large
lies between the loss of $G$ plus some measure of difference between $F$'s and $G$'s predictions
and the loss of $G$ plus another measure of difference between $F$'s and $G$'s predictions.

\subsection*{Acknowledgments}

I am grateful to the anonymous referees
\ifnotCONF
  of the conference version of this paper
\fi
for their comments.
This work was partially supported by MRC (grant S505/65).

\end{document}